\documentclass[conference,11pt]{IEEEtran}

\usepackage{graphicx}
\usepackage{fancyhdr}
\usepackage{pst-node,pst-text,pst-3d,pst-char}

\usepackage{hyperref}                       
\graphicspath{{figures/}}
\usepackage{tikz}							
\usepackage{pgfplots}						
\pgfplotsset{compat=newest}					
\usetikzlibrary{positioning}				
\usepackage{siunitx}						
\usepackage{booktabs}						
\usepackage{mathtools} 						
\usepackage{nicefrac} 						
\usepackage{multirow} 						
\usepackage{subcaption}						
\usepackage{cite}							
\usepackage[ruled]{algorithm2e}				
\usepackage{amssymb}						

\newcommand{\ego}{ego-vehicle }				
\newcommand{\egowo}{ego-vehicle}			
\newcommand{\tabelle}{Table~}				
\newcommand{\abbildung}{Fig.~}				
\newcommand{\gleichung}{Equation~}			
\renewcommand{\vec}[1]{\mathbf{#1}}			
\newcommand{\RDD}{RDD }						
\newcommand{\RDDwo}{RDD}					

\definecolor{TUMGreen}{RGB}{162, 173, 0} 
\definecolor{TUMOrange}{RGB}{227, 114, 34}
\definecolor{TUMGray}{RGB}{156, 157, 159}       
\definecolor{TUMGray3}{RGB}{217, 218, 219}       
\definecolor{TUMDiag1}{RGB}{105, 8, 90}  
\definecolor{TUMDiag14}{RGB}{249, 186, 0} 
\definecolor{TUMDiag9}{RGB}{0, 124, 48}       
\colorlet{green}{TUMGreen}  
\colorlet{red}{TUMOrange}

\newlength\figureheight             		
\newlength\figurewidth 						

\graphicspath{{figures/}}
\makeatletter
\def\input@path{{figures/}} 
\makeatother

\usetikzlibrary{external}			
\tikzset{external/system call={pdflatex -shell-escape -buf-size=10000000 -extra-mem-top=100000000 -halt-on-error -interaction=batchmode -jobname "\image" "\texsource"}} 

\usetikzlibrary{external}
\usepackage{textcomp}
\usepackage{lipsum}
\newcommand\copyrighttext{%
	\footnotesize \textcopyright 2020 IEEE.  Personal use of this material is permitted.  Permission from IEEE must be obtained for all other uses, in any current or future media, including reprinting/republishing this material for advertising or promotional purposes, creating new collective works, for resale or redistribution to servers or lists, or reuse of any copyrighted component of this work in other works. %
}
\newcommand\copyrightnotice{%
	\tikzset{external/export=false}
	\begin{tikzpicture}[remember picture,overlay]
	\node[anchor=south,yshift=-4pt, xshift=4pt] at (current page.south) {\fbox{\parbox
			{\dimexpr\textwidth-\fboxsep-\fboxrule\relax}{\copyrighttext}}};
	\end{tikzpicture}%
	\tikzset{external/export=true} %
}

\begin{document}
\title{{\large \vspace*{-3mm} \hspace*{-3mm} 2020 Fifteenth International Conference on Ecological Vehicles and Renewable Energies (EVER)}\vspace{5mm}\\
Identification of Challenging Highway-Scenarios for the Safety Validation of Automated Vehicles Based on Real Driving Data \vspace*{-7mm}}
%
%
%
\author{\IEEEauthorblockN{Thomas Ponn, Matthias Breitfu{\ss}, Xiao Yu and Frank Diermeyer}
\IEEEauthorblockA{Institute of Automotive Technology\\
Technical University of Munich\\
85748 Garching, Germany\\
Email: \textcolor[rgb]{0.00,0.07,1.00}{thomas.ponn@tum.de}} 
}
\maketitle %
\copyrightnotice %
\vspace{-0.6cm}
\begin{abstract}
For a successful market launch of automated vehicles (AVs), proof of their safety is essential. Due to the open parameter space, an infinite number of traffic situations can occur, which makes the proof of safety an unsolved problem. With the so-called scenario-based approach, all relevant test scenarios must be identified. This paper introduces an approach that finds particularly challenging scenarios from real driving data (\RDDwo) and assesses their difficulty using a novel metric. Starting from the highD data, scenarios are extracted using a hierarchical clustering approach and then assigned to one of nine pre-defined functional scenarios using rule-based classification. The special feature of the subsequent evaluation of the concrete scenarios is that it is independent of the performance of the test vehicle and therefore valid for all AVs. Previous evaluation metrics are often based on the criticality of the scenario, which is, however, dependent on the behavior of the test vehicle and is therefore only conditionally suitable for finding "good" test cases in advance. The results show that with this new approach a reduced number of particularly challenging test scenarios can be derived.
\end{abstract}


\IEEEpeerreviewmaketitle

\section{INTRODUCTION}
\label{sec:introduction}

For the validation and certification of automated vehicles (AVs; Level 3 and higher according to SAE \cite{SAEJ3016.2018}), the so-called scenario-based approach is a promising method for efficiently achieving a reliable safety statement about the AV. The scenario-based approach is intended to reduce the test effort by limiting the tests to meaningful scenarios. However, the question remains open as to how all necessary scenarios can be identified. The aim is to have a scenario catalog or database with scenarios that contains all "good" test cases for the safety assessment.

Scenarios can be generated either knowledge-based or data-based \cite{Riedmaier.2020}. The advantage of knowledge-based generation (e.\,g. using ontologies \cite{Bagschik.2018}) is that a comprehensive set of scenarios can be defined quickly and cost-effectively. However, the more time-consuming data-based approach is more promising for the creation of a complete set. 

\begin{figure}[b]
	\centering
	\includegraphics[width=0.95\columnwidth]{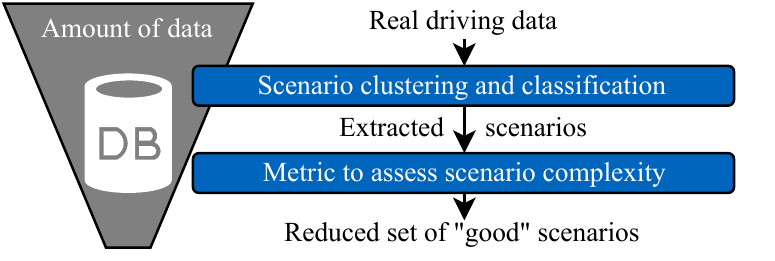}
	\caption{Data reduction during scenario extraction out of \RDDwo.}
	\label{fig:dataReduction}
\end{figure}

The data-based approach uses real driving data (\RDDwo; e.\,g. highD-dataset \cite{Krajewski.2018}) and extracts scenarios from it. First, individual scenarios are extracted from the large amount of data using clustering and classification methods, which is the first focus of this paper. In this initial step, as is the basic principle of the scenario-based approach, all free driving and similar uninteresting situations are omitted, thus reducing the amount of data (\abbildung \ref{fig:dataReduction}).

Subsequently, the extracted scenarios can be examined in more detail. One category of scenarios can be, for example, cut-in scenarios. Because this scenario category occurs frequently in \RDDwo, the measured data will contain a large number of cut-in situations. Not all of them should be stored in the database and then used as test cases, because many of them are redundant or contribute only minor to the safety assessment. Therefore, there are different metrics with which scenarios can be evaluated and thus a selection of particularly difficult (complex) test cases can be performed. The development of such a novel complexity-based metric is the second focus of this paper. As a result, a reduced number of particularly interesting (in relation to the metric used) scenarios is obtained for an efficient assessment of AVs. Therefore, the contributions of this paper are as follows:


%
\begin{itemize}
	\item Providing definitions for the differentiation of critical, challenging and complex scenarios (Section \ref{subsubsec:differentiation})
	\item Overall approach for the definition of a reduced set of 'good' scenarios based on complexity (Section \ref{subsec:overallApproach})
	\item Scenario clustering and classification based on \RDD (Section \ref{subsec:clusteringAndClassification})
	\item Novel and comprehensive metric for the assessment of scenarios regarding Layer 4 for the highway use-case which is the main contribution (Section \ref{subsec:complexityMetric})
\end{itemize}

\section{RELATED WORK}
\label{sec:related_work}
This section first defines important terms and then explains various methods for evaluating scenarios for AV testing.

\subsection{Terms and Definitions}
\label{subsec:termsAndDefinitions}

\textbf{Scenario}: In the context of this work, the definition of \textsc{Ulbrich et. al} \cite{Ulbrich.2015} is used, according to which a scenario is a temporal sequence of scenes, whereby actions and events of the elements involved occur within this sequence. By actions and events are meant, for example, maneuvers such as a cut-out situation or approaching the end of a traffic jam. On this basis, \textsc{Menzel et. al} \cite{Menzel.2018} define three different categories of scenarios. These are the so-called functional, logical and concrete scenarios. Starting with a verbal description of the functional scenarios, through the logical scenarios defined by parameter ranges and distributions, to the concrete scenarios defined by exact parameter values, the level of detail and machine readability increases. For logical and concrete scenarios, all parameters that describe the scenario are required. For this purpose, a five-layer model for structuring the parameters is presented in \cite{Bagschik.2018}. The five layers are defined as below:
\begin{itemize}
	\item[] Layer 1: Road-level
	\item[] Layer 2: Traffic infrastructure
	\item[] Layer 3: Temporary manipulation of L1 and L2
	\item[] Layer 4: Objects
	\item[] Layer 5: Environment
\end{itemize} 


\textbf{Traffic Participants (TP)}: In general, all movable objects, such as pedestrians and cyclists are TPs. Due to highways as the considered use case represented by the used \RDDwo, the term traffic participant is used in this paper as a synonym for passenger cars and trucks.

\begin{figure}[b]
	\centering
	\includegraphics[width=0.95\columnwidth]{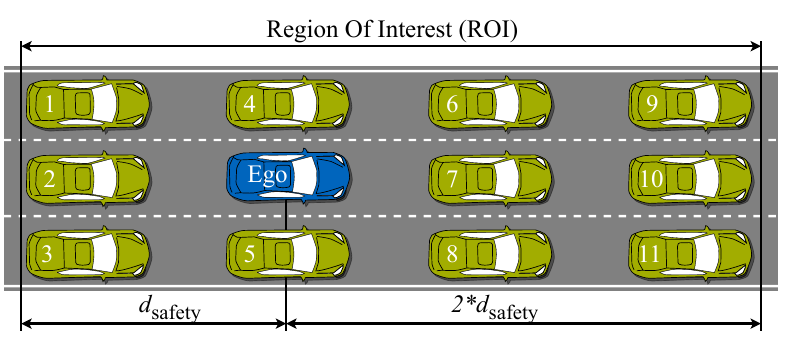}
	\caption{Definition of Region Of Interest (ROI) for highway scenarios based on eleven surrounding TP (adapted from \cite{AntonaMakoshi.2019}) and the safety distance $d_{\text{safety}}$ defined as a time gap of \SI{1.8}{\second}.}
	\label{fig:ROI}
\end{figure}

\textbf{Region of Interest (ROI)}: For the extraction of scenarios from \RDD all relevant data about the environment of the \ego have to be considered. In this paper we focus on the objects from Layer 4 according to the five-layer model of \cite{Bagschik.2018} and consider the surrounding traffic participants analogous to \cite{AntonaMakoshi.2019} as relevant when within the ROI of the \egowo. The definition of the ROI is visualized in \abbildung \ref{fig:ROI}.

\subsection{Scenario Selection Methods}
\label{subsec:scenarioAssessmentMethods}

This section first presents selection approaches that describe if a scenario is critical and/or challenging. It becomes apparent that there is no common understanding in the literature about the definition of these terms. Therefore, a clear distinction of the terms is proposed at the end of this section.

\subsubsection{Selection of Critical Scenarios}
\label{subsubsec:CriticalityBased}

The use of a criticality metric for scenario selection is a commonly applied approach. The best known metric is the time-to-collision (TTC) \cite{Hayward.1972}. 
An overview of criticality metrics can also be found in \cite{Mahmud.2017}. If the used \RDD were recorded by human drivers, the problem with criticality-based selection of scenarios is that the same traffic situations do not necessarily have to be as critical for AVs as for human drivers.

\textsc{Hallerbach et al.} \cite{Hallerbach.2018} define four types of criticality: individual, nanoscopic, microscopic and macroscopic. This is used to evaluate criticality in different spatial areas around the AV. They use a learning-based procedure to make an overall binary decision based on the four criticality areas as to whether the scenario is critical or not. The generation of critical scenarios based on \RDD is the focus of \cite{Klischat.2019, Althoff.2018}. The criticality of a scenario is calculated based on the size of the area that can be safely used by the AV. Optimization by means of evolutionary algorithms maximizes the criticality of the scenarios by adapting the behavior of surrounding TPs and minimizes the area that can be safely used. In \cite{Pierson.2019}, a method is introduced that allows the risk of \RDD to be efficiently determined in order to select critical scenarios for testing AVs. The risk in a spatial location relates to the position and speed of surrounding traffic participants.

\subsubsection{Selection of Challenging Scenarios}
\label{subsubsec:ComplexityBased}

The use of the Analytic Hierarchy Process to identify challenging scenarios is investigated by \cite{Gao.2019, Xia.2018, Xia.2017}. \textsc{Wang et al.} \cite{Wang.2018} use a two-stage assessment to determine if the static environment and the dynamic surrounding is challenging. \textsc{Qi et al.} \cite{Qi.2019} use the so called Scenario Character Parameter (SCP) based on the trajectories, which lead to an insufficient \ego performance. By analyzing the SCP, scenario groups can be created and reduced to one challenging scenario. An optimization-based approach (without concrete implementation) for defining challenging scenarios is presented in \cite{Ponn.2019c}. Aspects of this method are examined in more detail in \cite{Ponn.2019b,Ponn.2019d}. \textsc{Bolte et al.} \cite{Bolte.2019} extract challenging scenarios from data based on the difficulty of predicting the future behavior of surrounding TPs.

Compared to criticality-based selection, there is little literature available for identifying challenging scenarios. It is noticeable, however, that the terms challenging and complex are often mixed up and used as synonyms. In rare cases it is also called a corner case. What all three terms have in common is that they describe a scenario that is particularly difficult for the AV.

\subsubsection{Differentiation Between Critical and Challenging Scenarios}
\label{subsubsec:differentiation}

The terms critical, challenging and complex are not used consistently in literature. In our understanding, challenging and complex cannot be separated from each other clearly. However, it is possible to clearly separate these terms from criticality. The separation is done according to the purpose of the evaluation and whether it can be performed before or after the test case execution. If criticality is evaluated, the behavior of the \ego is assessed in a concrete scenario. This evaluation can only be made after execution of the test case. If, on the other hand, the concrete scenario itself is to be evaluated, it can be classified as challenging/complex. According to this, the scenarios of \cite{Klischat.2019, Althoff.2018} should be assigned to challenging even when the authors use the term critical. In the context of our work, we also consider challenging as an umbrella term for complex scenarios. While the difficulty of challenging scenarios can exist in any parameter layer, complex scenarios describe particularly difficult ones in relation to Layer 4 (\abbildung \ref{fig:complexCritical}). The contribution of the present paper can thus be assigned to complexity. Therefore, we use the following definitions: 

\begin{figure}[t]
	\centering
	\includegraphics[width=0.95\columnwidth]{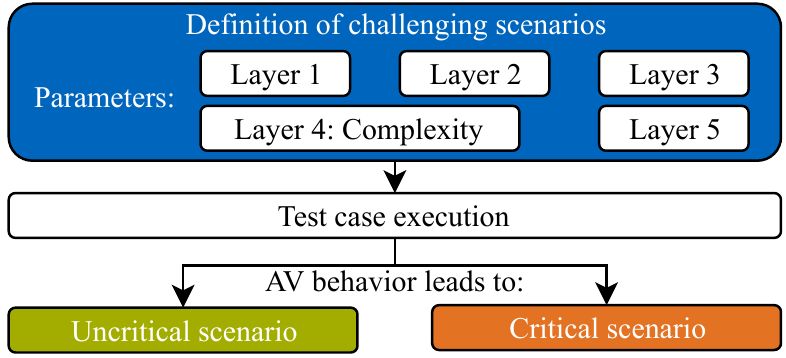}
	\caption{Difference between challenging, complex and critical scenarios. Complexity means the difficulty of a scenario due to the behavior (trajectories) of the objects (TPs) that are part of Layer 4 of the five-layer model of \cite{Bagschik.2018}.}
	\label{fig:complexCritical}
\end{figure}

\textbf{Critical}: Assessment of the performance of the \ego behavior in a concrete scenario. It is only determinable after test case execution and the behavior of different AV-functions lead to different criticality-results for the same concrete scenario.

\textbf{Challenging and Complex}: It means an assessment of a concrete scenario itself. It is determinable before test case execution and independent of the AV-performance. Whether a concrete scenario is challenging / complex or not, depends on the chosen parameter values. Therefore, challenging or complex can be seen as the difficulty for the AV to master the concrete scenario without the occurrence of a critical situation. The main assumption herein is that more complex scenarios lead more often to critical situations when they are executed.

\section{METHODOLOGY}
\label{sec:methodology}
This section first gives a brief overview of the overall method and then explains in detail the approach of the scenario clustering and classification as well as the metric for complexity assessment. 

\subsection{Overall Approach}
\label{subsec:overallApproach}

\begin{figure}[b]
	\centering
	\includegraphics[width=0.95\columnwidth]{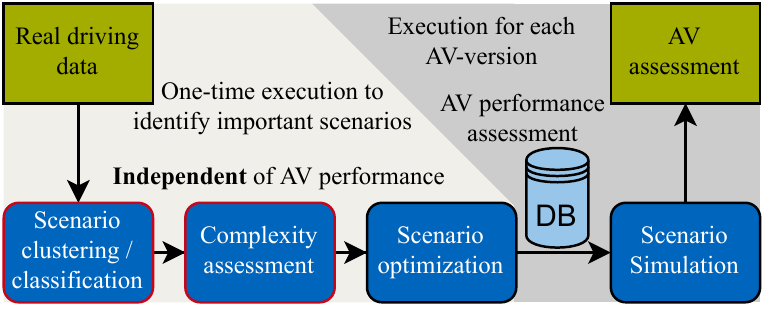}
	\caption{Approach to identify a reduced set of challenging scenarios for the test of AVs. The focus of the present paper is the scenario clustering / classification and the complexity assessment (highlighted in red).}
	\label{fig:overalApproach}
\end{figure}

The overall method is a procedure to identify a reduced set of particularly important scenarios based on \RDDwo, with which various automated driving functions can be tested. A visualization as well as the highlighting of the sub-methods discussed in this paper can be seen in \abbildung \ref{fig:overalApproach}. 


The scenario clustering and classification method as well as the complexity assessment are explained in detail in Sections \ref{subsec:clusteringAndClassification} and \ref{subsec:complexityMetric}. Scenario optimization, simulation and the assessment of the AV are not covered in this paper and are only briefly explained here for the sake of completeness. In scenario optimization, the complexity of the scenarios extracted from the \RDD is further increased in order to improve the quality of the scenarios. This is achieved by adapting the behavior of the surrounding TPs. The optimized scenarios can then be stored in a database. In scenario simulation, an actual driving function is used for the first time in the methodology. Its performance in the optimized scenarios is then examined and evaluated using, for example, criticality-based key performance indicators. 

\subsection{Scenario Clustering and Classification}
\label{subsec:clusteringAndClassification}

\begin{figure}[b]
	\centering
	\includegraphics[width=0.95\columnwidth]{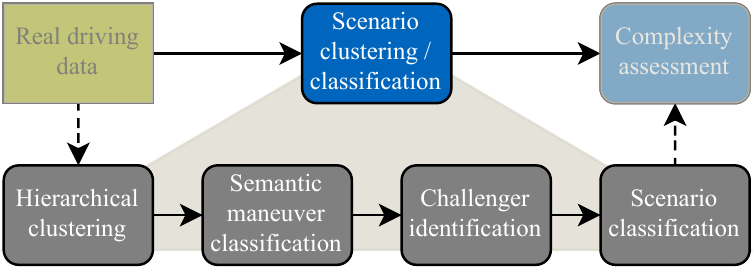}
	\caption{Sub-methods of the scenario clustering and classification block of \abbildung \ref{fig:overalApproach}.}
	\label{fig:clusteringApproach}
\end{figure}

The basis for clustering and classifying scenarios are \RDDwo. In this paper, we use the highD data set \cite{Krajewski.2018}. The developed process consists of four steps, that are shown in \abbildung \ref{fig:clusteringApproach}. 

\subsubsection{Hierarchical Clustering}
\label{subsub:hierarchicalClustering}
The highD data consists of a \SI{420}{\meter} long highway section recorded by a drone. The vehicles need in median \SI{13.6}{\second} to pass through this section. Hierarchical clustering is used in combination with the ROI defined in Section \ref{subsec:termsAndDefinitions} to extract individual scenarios from the several minutes of recorded data. Each cluster consists of an \ego and all relevant surrounding TPs. Thereby, uninteresting free driving situations (no surrounding TPs exist) are sorted out to reduce the amount of data. 

\subsubsection{Semantic Maneuver Classification}
We use a rule-based approach for maneuver classification as all driving situations in highway traffic can be distinguished with a manageable amount of rules. Furthermore as \cite{Erdogan.2019} claims, the rule-based performance is almost as good as with learning-based approaches. For each TP in the scenario, a rule-based decision tree is used to define the maneuver with respect to the \egowo. The correct maneuver is selected from 13 different maneuvers, such as 'overtaking', 'following drive of the \egowo' or 'driving parallel to the \egowo'. 

\subsubsection{Challenger Identification}
\label{subsub:challengerIdentification}

Challengers are defined as TPs that force reactions by the \egowo, in order to avoid potential collisions. The determination of those vehicles is accomplished by predicting the trajectories of the \ego and analyzing potential intersections with other TPs. A scenario can have none, one or several challengers. Scenarios without a challenger can be sorted out because no reaction of the \ego is necessary and therefore no contribution to the safety assessment is made. For scenarios with a Challenger all other TPs are also relevant, as they can represent action restrictions for the \egowo. For scenarios with more than one challenger, the first challenger is considered decisive for the following scenario classification. 

\subsubsection{Scenario Classification}
\label{subsub:scenarioClassification}

\begin{figure}[t]
	\centering
	\includegraphics[width=0.95\columnwidth]{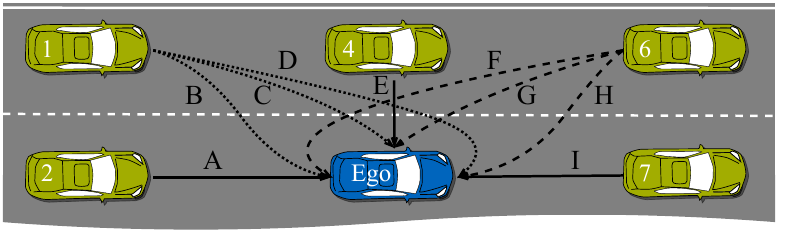}
	\caption{Scenario classification based on the challenger trajectory. Only the left side is shown from the perspective of the \egowo, because the trajectories and designations are symmetrical. The figure is based on \cite{Weber.2019}.}
	\label{fig:functionalScenarios}
\end{figure}

Based on the trajectory of the challenger, the scenario is assigned to one of nine functional scenarios used in the PEGASUS project \cite{Weber.2019} (\abbildung \ref{fig:functionalScenarios}). The maneuver classification in step 2) uses more than nine categories, because the behavior of all TPs (including non challenger) is classified.

In summary, concrete scenarios assigned to one of nine functional scenarios based on the trajectory of the challenger represent the result of the clustering and classification method. In addition, the amount of data is reduced by neglecting scenarios without challengers.

\subsection{Definition of the Complexity Metric}
\label{subsec:complexityMetric}

To further reduce the number of scenarios and thus the amount of data, the classified scenarios are evaluated using a novel complexity metric. This allows us to sort out uninteresting scenarios and to focus on highly complex scenarios that are more difficult for the AV. Under the assumption made in Section \ref{subsubsec:differentiation}, this leads to a higher probability of the occurrence of erroneous behavior of the AV. 

The developed complexity metric aims at evaluating the difficulty of the scenario itself without considering \egowo's behavior. Thereby the following assumptions are made: 

\begin{itemize}
	\item There are different attributes of a traffic situation that contribute to complexity,
	\item All attributes are calculated for all TPs within the ROI,
	\item Different attributes have different importance,
	\item Coupling effects between attributes are neglected,
	\item Every attribute has linear contribution to the overall complexity,
	\item The maximum complexity during the scenario is used as the descriptive value.
\end{itemize}

The scalar complexity of a scenario $C_{\text{scenario}}$ can then be calculated according to \gleichung \ref{eq:C}. 
\begin{equation}
C_{\text{scenario}} = \max(\vec{w}^T \cdot \vec{a}_i) \quad \text{with} \quad i \in \{1,...,m\} 
\label{eq:C}
\end{equation}
Hereby $\vec{w} \in (n\times1)$ is the weighting vector, $\vec{a}_i \in (n\times1)$ the attribute vector, $m$ is the number of scenes in the scenario and $n$ is the number of attributes. The sum of all $n$ weights $w_j$ and the normalization of all attributes $a_{i,j}$ is shown in \gleichung \ref{eq:weights}, where $a_{i,j}$ denotes the $j$-th entry of the attribute vector in the $i$-th scene of the scenario under consideration.  
\begin{equation}
\sum_{j=1}^{n} w_j = 1 \quad \text{and} \quad a_{i,j} \in [0,1] 
\label{eq:weights}
\end{equation}

The number of scenes in a scenario $m$ can be calculated using \gleichung \ref{eq:tScenario}, where $t_{\text{scenario}}$ is the time duration of the scenario. The time step size $\Delta t$ in the highD data is specified as \SI{0.04}{\second} since the camera of the drone has a frequency of \SI{25}{\hertz}. With \SI{13.6}{\second} as median appearance of the vehicles, the median number of scenes $m$ is \SI{340}{}.
\begin{equation}
m = \frac{t_{\text{scenario}}}{\Delta t} 
\label{eq:tScenario}
\end{equation}
The choice of attributes $a_j$ has a crucial influence on the calculation of the complexity. A complete list of influencing factors cannot be determined objectively, therefore $n=13$ different attributes are defined based on literature as well as expert knowledge from industry and research. A brief summary is given below and a more comprehensive description can be found in our preliminary investigation in \cite{Yu.2019}.
\begin{enumerate}
	\item Number of surrounding TPs
	\item Types (car, truck, ...) of surrounding TPs
	\item Dynamic (velocity, acceleration) of surrounding TPs
	\item Variation of dynamic parameters of surrounding TPs
	\item Number of action dependencies of surrounding TPs
	\item Predictability (with simple constant acceleration model) of future behavior of surrounding TPs 
	\item Time-gap between \ego and surrounding TPs
	\item Possible actions of \ego (due to action restriction by other TPs)
	\item Possible actions of surrounding TPs (due to action restriction by other TPs)
	\item Occluded area for \ego due to line-of-sight obstruction by other TPs
	\item Number of actions performed by surrounding TPs during the scenario
	\item Time to brake for \ego to avoid accidents
	\item Number of actions of \ego during the scenario
\end{enumerate}

All attributes that cannot be calculated in every scene, such as the number of actions of individual TPs, are calculated at the end of the scenario and their values are used for all scenes. This ensures a consistent comparison between different scenarios. To also make the influence of the attributes on the overall complexity comparable, all attributes are normalized to a value range from 0 to 1 according to \gleichung \ref{eq:weights}. 

\section{RESULTS}
\label{sec:results}

This section shows the results of the scenario classification and complexity assessment. Finally, the complexity metric is validated.

\subsection{Scenario Clustering and Classification}
\label{subsec:resultsClusteringAndClassification}

\begin{figure}[]
	\centering
	\begin{tikzpicture}
\definecolor{mycolor1}{rgb}{0.00000,0.39608,0.74118}%
\begin{axis}[
    ybar,
    width=8.0cm,
    height=4cm,
    legend style={at={(0.5,-0.15)},
      anchor=north,legend columns=-1},
    ylabel={Number of scenarios},
    symbolic x coords={A,B,C,D,E,F,G,H,I},
    xtick=data,
    xtick pos=bottom,
    xlabel={Functional scenario},
    ymin=0,
    ]
\addplot[color=black, fill=mycolor1] coordinates {(A,12449) (B,1233) (C,6009) (D,403) (E,2336) (F,137) (G,6620) (H,1966) (I,20090)};
\end{axis}
\end{tikzpicture}
	\caption{Number of concrete scenarios in the highD dataset which include a challenger. They are assigned to the nine defined functional scenarios from \abbildung \ref{fig:functionalScenarios}.}
	\label{fig:numberOfScenarios}
\end{figure}
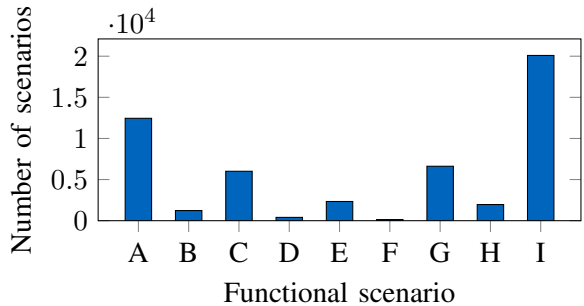

The highD data set contains 110,507 vehicles, which means that theoretically 110,507 concrete scenarios can be extracted. As described in Section \ref{subsub:hierarchicalClustering}, clustering already eliminates all scenarios without surrounding TPs. After this step, 110,007 scenarios remain from the highD data for further processing. The number of relevant scenarios can be further reduced to 67,455 by the challenger consideration (Section \ref{subsub:challengerIdentification}). These 67,455 concrete scenarios are assigned to the nine defined functional scenarios using the classification method from Section \ref{subsub:scenarioClassification}. The distribution of the functional scenarios can be seen in \abbildung \ref{fig:numberOfScenarios}. 

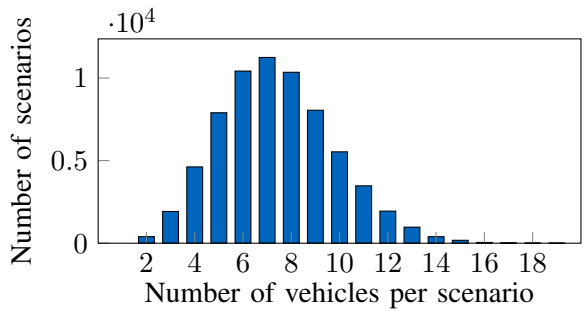
\begin{figure}[]
	\centering
	\begin{tikzpicture}
\definecolor{mycolor1}{rgb}{0.00000,0.39608,0.74118}%
\definecolor{TUMIvory}{rgb}{0.00000,0.39608,0.74118}%
\begin{axis}[
    hist,
    width=8.0cm,
    height=4.3cm,
    set layers,
    axis on top,
	x label style={at={(axis description cs:0.5,-0.14)},anchor=north},
    ylabel={Number of scenarios},
    xlabel={Number of vehicles per scenario},
    xtick pos=bottom,
    xtick={2,4,6,8,10,12,14,16,18},
    ytick pos=left,
    ymin=0,
    xmin=0.0,
    xmax=20.0,
    ] 
\addplot[ybar,bar width=6pt,fill=mycolor1,opacity=1.0] 
table[col sep=comma] {figures/vehiclesPerScenario.txt}; 
\end{axis}
\end{tikzpicture}
	\caption{Distribution of the number of involved vehicles per scenario, whereby the ego vehicle is also counted. The median is seven and the maximum is 19 vehicles.}
	\label{fig:vehiclesPerScenario}
\end{figure}

Figure \ref{fig:vehiclesPerScenario} shows the number of vehicles involved in the scenarios. The \ego is also included, so the minimum number of vehicles is two. In these scenarios only the challenger is present in addition to the \egowo. These types of scenarios only account for \SI{0.59}{\percent} of the total scenarios. In all other scenarios there are additional TPs that have to be considered when planning a safe trajectory and which can also represent action restrictions for the \egowo. 


\subsection{Complexity Assessment}
\label{subsec:compleA comparisxityAssessment}

In order to calculate the complexity, the weights $w_j$ of all attributes were determined by an online expert survey. Using a two-part online survey, 25 experts from the field of safety assessment of AVs were first asked how important they consider each of the $n$ attributes in terms of complexity for AVs. Then, in a second part, a subgroup of 20 experts were shown 20 scenarios of the highD dataset on video\footnote{The videos are available via \url{https://www.youtube.com/channel/UC3IV32GfmVKXouqvF74jMFg/videos?view=0&sort=dd&shelf_id=0}}, which they evaluated from low to high complexity. The weights were then determined by means of a compromise in such a way that both the expert opinion on the importance of each attribute as well as the evaluation of the scenarios were represented by the developed metric. This compromise is necessary because the results of the two parts of the online survey do not match exactly, i.\,e. if high rated attributes of the experts are weighted too high, the metric reflects the expert opinion worse in the evaluation of the 20 scenarios. The resulting weighting vector $w_j$ is

\begin{equation}
\begin{aligned}
w = &[0.01, 0.087, 0.087, 0.1, 0.087, 0.077, 0.087, \\
    & 0.087, 0.087, 0.087, 0.1, 0.02, 0.084].
\label{eq:finalWeighting}
\end{aligned}
\end{equation}

Using the determined weighting factors, all challenger scenarios are evaluated with the complexity metric. 

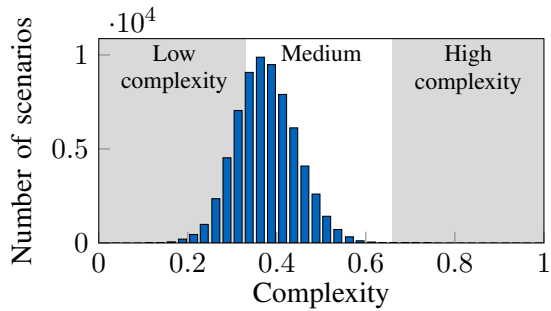
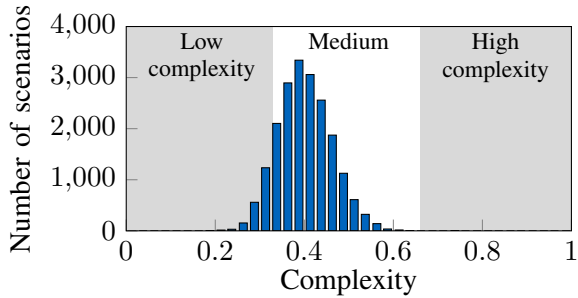
\begin{figure}[]
	\centering
	\begin{subfigure}[t]{.45\textwidth}
		\centering
		\begin{tikzpicture}
\definecolor{mycolor1}{rgb}{0.00000,0.39608,0.74118}%
\definecolor{TUMIvory}{rgb}{0.00000,0.39608,0.74118}%
\begin{axis}[
    hist,
    width=7.5cm,
    height=4.3cm,
    set layers,
    axis on top,
	x label style={at={(axis description cs:0.5,-0.14)},anchor=north},
    ylabel={Number of scenarios},
    xlabel={Complexity},
    xtick pos=bottom,
    ytick pos=left,
    ymin=0,
    xmin=0.0,
    xmax=1.0,
    ] 
\filldraw [fill=gray, draw=none, opacity=0.3] (rel axis cs:0,0) rectangle (rel axis cs:0.33,1);
\filldraw [fill=gray, draw=none, opacity=0.3] (rel axis cs:0.66,0) rectangle (rel axis cs:1,1);
\addplot[ybar,bar width=3pt,fill=mycolor1,opacity=1.0] 
table[col sep=comma] {figures/allScenarioComplexityData.txt}; 
\node[align=center,anchor=north,text width=1.5cm] at (rel axis cs:0.17,1.02) {\footnotesize Low\\[-1mm] complexity};
\node[align=center,anchor=north,text width=1.5cm] at (rel axis cs:0.50,1.02) {\footnotesize Medium};
\node[align=center,anchor=north,text width=1.5cm] at (rel axis cs:0.83,1.02) {\footnotesize High\\[-1mm] complexity};
\end{axis}
\end{tikzpicture}
		\caption{Complexity distribution of all challenger scenarios. Only 8 out of the 67,455 challenger scenarios are classified with high complexity.}
		\label{fig:scenarioComplexityAllClasses}
	\end{subfigure}%
	\hfill
	\begin{subfigure}[t]{.45\textwidth}
		\centering
		\begin{tikzpicture}
\definecolor{mycolor1}{rgb}{0.00000,0.39608,0.74118}%
\definecolor{TUMIvory}{rgb}{0.00000,0.39608,0.74118}%
\begin{axis}[
    hist,
    width=7.5cm,
    height=4.3cm,
    set layers,
    axis on top,
	x label style={at={(axis description cs:0.5,-0.14)},anchor=north},
    ylabel={Number of scenarios},
    xlabel={Complexity},
    xtick pos=bottom,
    ytick pos=left,
    ymin=0,
    ymax=4000,
    xmin=0.0,
    xmax=1.0,
    ] 
\filldraw [fill=gray, draw=none, opacity=0.3] (rel axis cs:0,0) rectangle (rel axis cs:0.33,1);
\filldraw [fill=gray, draw=none, opacity=0.3] (rel axis cs:0.66,0) rectangle (rel axis cs:1,1);
\addplot[ybar,bar width=3pt,fill=mycolor1,opacity=1.0] 
table[col sep=comma] {figures/scenarioIComplexityData.txt}; 
\node[align=center,anchor=north,text width=1.5cm] at (rel axis cs:0.17,1.02) {\footnotesize Low\\[-1mm] complexity};
\node[align=center,anchor=north,text width=1.5cm] at (rel axis cs:0.50,1.02) {\footnotesize Medium};
\node[align=center,anchor=north,text width=1.5cm] at (rel axis cs:0.83,1.02) {\footnotesize High\\[-1mm] complexity};
\end{axis}
\end{tikzpicture}
		\caption{Distribution of the scenario complexity of the 20,090 concrete scenarios of functional scenario I.}
		\label{fig:scenarioIComplexity}
	\end{subfigure}
	\caption{Complexity distribution of the extracted highD scenarios using an equidistant classification into low, medium and high complexity. Each scenario class shows a similar distribution compared to the distribution of all scenarios.}
	\label{fig:auxiliary2}
\end{figure}

Figure \ref{fig:scenarioComplexityAllClasses} shows the complexity distribution for all scenarios. Additionally, \abbildung \ref{fig:scenarioIComplexity} depicts the distribution for the functional scenario I. The distribution for scenario class I (as well as all other functional scenarios) is similar to the distribution for all scenarios. This means that no conclusions can be drawn about complexity based on the scenario class. 

\begin{figure*}[]
	\centering
	    \begin{subfigure}[]{.95\textwidth}
	    	\centering
	    	\includegraphics[width=\columnwidth]{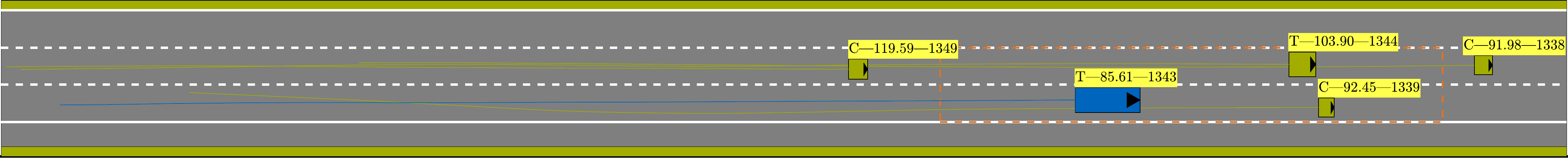}
	    	\caption{Scenario with low complexity $C_{\text{scenario}}=0.32$.}\label{fig:1a}
	    \end{subfigure} %
		\par\bigskip
 	    \begin{subfigure}[]{.95\textwidth}
    	    	\centering
    	    	\includegraphics[width=\columnwidth]{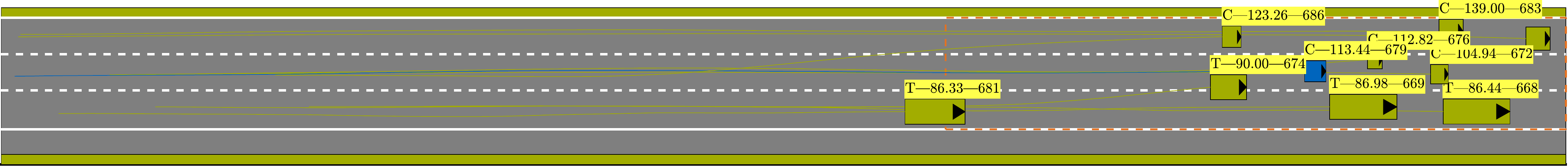}
    	    	\caption{Scenario with high complexity $C_{\text{scenario}}=0.66$.}\label{fig:1b}
 	    \end{subfigure} %
	\caption{Comparison of scenarios with low and high complexity. The \ego including the driven path is shown in blue and all surrounding TPs are shown in green. The ROI of the \ego at the depicted time step is marked by a dashed orange rectangle. Each vehicle has a three-part identification consisting of the class (car or truck), the current speed in km/h and the highD vehicle ID.}
	\label{fig:scenarioComparison}
\end{figure*}

Due to space limitations, the complexity distribution of only one functional scenario (Scenario I) is shown in \abbildung \ref{fig:scenarioIComplexity}. It is noticeable that a large number of scenarios from \RDD have only a low and medium complexity, respectively. Given the complexity levels used, if only scenarios with high complexity are considered for the further process, this corresponds to an additional reduction of \SI{99.99}{\percent}. A comparison between a low complexity scenario and a high complexity scenario is shown in \abbildung \ref{fig:scenarioComparison}. It can be seen that in the scenario with high complexity, it is more difficult for the AV to plan a safe trajectory.  



It is interesting to compare our approach with a criticality-based selection of scenarios. In \cite{Pierson.2019}, vehicle 1034 from highD track 25 is used as an example for a critical scenario. Exactly this scenario is assigned to the functional scenario~I in our framework and evaluated with only a moderate complexity of 0.37. The reason for this is the low speed of approximately 20 km/h and the sufficient space for an evasive maneuver. So the scenario itself is not very difficult, but the performance of the human driver was not optimal, so that a critical situation occurred. Including this test case in the database therefore offers no benefit if a reliable database for the evaluation of different AVs is to be established.

\subsection{Validation}
\label{subsec:validation}

One aspect of validation is the completeness of the attributes. This was evaluated in the online expert survey. When assessing the importance of the individual attributes, the experts were asked to recommend changes in the implementation of the attribute. At the end of the 13 attributes, the experts were asked for additional attributes that they deemed should be taken into account. For the former, only minor changes were suggested by the experts and for the latter, no new attribute was named by the experts. Thus the complexity metric can be considered complete.  

The second aspect is the validation of the statement from Section \ref{subsubsec:differentiation} that more complex scenarios lead more often to critical situations. For this purpose, a simple automated driving function is developed in Matlab/Simulink using the Automated Driving Toolbox. Subsequently, scenarios from the highD data set are simulated, which can be seen as a simplified implementation of the block 'Scenario Simulation' from Figure \ref{fig:overalApproach}. The simple driving function thereby substitutes the \ego from the original highD scenario and consists of a combination of existing examples from the Automated Driving Toolbox, which include ACC, LKA and an emergency brake assistant. In addition, a lane change function is implemented so that the vehicle can change lanes when the adjacent lane is free. Since all highD data were recorded on highways, these four sub-functions are sufficient to enable a vehicle to drive automatically in the considered highway use case.   

The scenarios from the highD data set are created using the Scenario Designer of the Automated Driving Toolbox. Because the complexity metric is not completely independent of the behavior of the \egowo, the scenarios are started at the time with the highest complexity of the scenario. The behavior of the surrounding road users is predefined and not adapted to the actions of the \ego (the AV to be tested). Therefore accidents can occur where the \ego is not to blame. Examples are situations in which a TP drives into the rear of the \ego because the AV drives slower than the original \ego from the highD data set. These scenarios are not considered in the evaluation. 

To prove the assumption that more complex scenarios lead more often to critical situations, scenarios with low, medium and high complexity are simulated, evaluated and the criticality is compared on the basis of the minimum TTC that occurred. Since there are only a few scenarios with high complexity in the entire data set, about one percent (650 scenarios) of the 67,455 scenarios with the lowest and highest complexity rating are selected respectively. In addition, 650 scenarios are selected that are closest to the average complexity of 0.38. This allows the three categories of lowest, average and highest complexity to be compared. 

\begin{figure}[]
	\centering
	\pgfplotsset{compat=newest}
\pgfplotsset{%
	xlabel style={at={(axis description cs:0.5,-0.13)}}, 
	ylabel style={at={(axis description cs:-0.10,0.5)}},
	y tick label style={/pgf/number format/fixed}
}
\definecolor{mycolor1}{rgb}{0.63529,0.67843,0.00000}%
\definecolor{mycolor2}{rgb}{0.00000,0.39608,0.74118}%
\definecolor{mycolor3}{rgb}{0.89020,0.44706,0.13333}%
\begin{tikzpicture}

\footnotesize
	\clip(-4,-2.2)rectangle(4.5,2.5);
	\node at (0,0) {
	\setlength\figureheight{3.5cm} 
	\setlength\figurewidth{7cm}
	\input{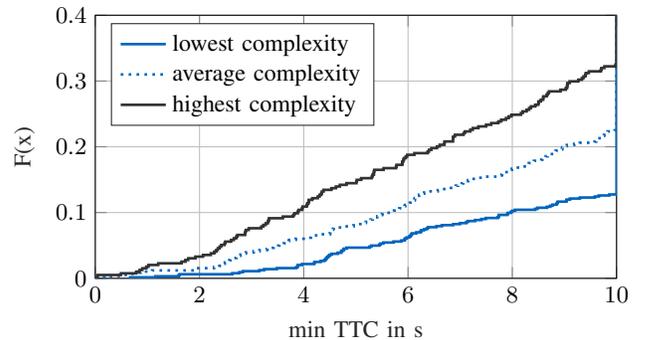} 
	};

\end{tikzpicture}%
	\caption{Cumulative distribution functions of the TTC of the lowest, average and highest rated scenarios. The three categories each contain approximately one percent of the scenarios shown in Figure \ref{fig:scenarioComplexityAllClasses}. It can be seen that scenarios with higher complexity more often show small TTC values. This means that more complex scenarios are on average more critical. }
	\label{fig:TTCcdf}
\end{figure}

 The evaluation (\tabelle \ref{tab:results}) shows that of the 650 scenarios per class, different numbers of scenarios remain after sorting out the irrelevant accidents. It can be seen that accidents in which the AV is not to blame occur more frequently with higher complexity. Despite the slightly lower number of test scenarios, accidents caused by the AV occur most frequently in highly complex scenarios and most frequently fall below the critical TTC value of 1.5 seconds. The critical TTC value is based on \cite{vanderHorst.1993}. In the original highD data set no accidents occur, so it can be concluded that the performance of the simple AV in these scenarios is worse than the human driver. This is plausible because it is a very simple system. In addition, \abbildung \ref{fig:TTCcdf} shows the cumulative distribution function of the minimum TTC occurring in the scenarios. Here again, it is confirmed that the most complex scenarios more often have low TTC values and are therefore more critical. It should be noted that the maximum value of the minimum TTC is limited to 10 seconds if the minimum TTC value is not below this value in the scenario. On the basis of these results, the assumption that more complex scenarios lead more often to critical situations can be confirmed and thus the functionality of the metric is proven.

\begin{table}[]
	\centering
	\caption{Results of the Matlab/Simulink simulation. The number of scenarios is not identical because all accidents that are not caused by the AV are sorted out. The critical TTC value is 1.5 seconds and is based on \cite{vanderHorst.1993}.}
	\label{tab:results}
	\begin{tabular}{lccc}
		\toprule
		& \multicolumn{3}{c}{Complexity class} \\
		& Lowest     & Average    & Highest    \\
		\midrule
		Number of Scenarios    			& 645        & 595        & 416      \\
		Scenarios below critical TTC 	& 2          & 7          & 9         \\
		Number of accidents    			& 2          & 13         & 22         \\
		\bottomrule
	\end{tabular}%
\end{table}

\section{DISCUSSION}
\label{sec:discussion}

The most significant influence on the quality of the results is the complexity metric being used. The basic idea is to evaluate only the scenario itself without considering the behavior of the \ego (which is often a human driver in case of currently available \RDDwo). This is not \SI{100}{\percent} possible because a scenario is a dynamic sequence of scenes where the \ego behavior influences the behavior of the surrounding TP and vice versa. Nevertheless, the evaluation is much less dependent on the behavior of the \ego than with the criticality-based selection of scenarios.

With the developed methodology it has been shown, that a reduced set of particularly challenging scenarios can be derived, at least with respect to Layer 4 of the five-level model of \cite{Bagschik.2018}. This can be very helpful especially for spot checking during certification of AV conducted by independent third parties. However, a further research question arises, which share of scenarios should be further processed to scenario optimization. Thereby, a distance measure can be used to measure the uniqueness of the scenario in order not to store too many similar scenarios in the database and at the same time not lose too much in completeness. In addition, individual functional scenarios can be examined more intensively in future work and an individual selection of scenarios can be made. 

During the simulation of the scenarios, accidents occur for which the AV is not to blame. This can be analyzed and eliminated in further work by specifically adapting the behavior of the surrounding TP in such cases. A more sophisticated driving function can also be used to achieve a performance in the scenario evaluations that is more comparable with that of the human driver. 

Although this approach is a promising method for identifying relevant scenarios, further sources for filling the scenario database will be necessary for a comprehensive safety assessment. Especially after small changes to the driving function, it can be advantageous to focus on critical scenarios of the previous version. 

\section{CONCLUSION}
\label{sec:conclusion}


This contribution addresses a novel method for complexity-based identification of a reduced set of especially challenging scenarios from real driving data. In contrast to commonly used criticality metrics, which evaluate the performance of the \ego behavior within a scenario, the developed complexity metric evaluates the scenario itself and can derive a generally valid catalog of 'good' test scenarios. Already during the clustering and classification of the scenarios about \SI{39}{\percent} of the data can be classified as uninteresting based on a challenger consideration and thus be sorted out. Using an equidistant classification into low, medium and high complexity, the most challenging scenarios can be identified and used for the further processing steps, allowing a further reduction of the data amount. The results proof the validity of the metric and therefore, the methodology presented here can contribute significantly to an efficient safety assessment of automated vehicles.

\section*{ACKNOWLEDGMENT AND CONTRIBUTIONS}

Thomas Ponn (corresponding author) initiated and wrote this paper. He was involved in all stages of development and primarily developed the research question as well as the concept. Matthias Breitfu{\ss} wrote his thesis on scenario clustering and classification and implemented the clustering and classification algorithms during his thesis. Xiao Yu wrote her thesis on the development of a complexity metric and implemented the complexity assessment algorithms during her thesis. Frank Diermeyer contributed to the conception of the research project and revised the paper critically for important intellectual content. He gave final approval of the version to be published and agrees to all aspects of the work. As a guarantor, he accepts responsibility for the overall integrity of the paper.

The research project was funded and supported by T\"UV S\"UD Auto Service GmbH.

\bibliographystyle{./bibliography/IEEEtran} 
\bibliography{./bibliography/IEEEabrv,./bibliography/referencesITSC}

\end{document}